\documentclass[11pt]{article}

\usepackage{acl}

\usepackage{times}
\usepackage{latexsym}
\usepackage[T1]{fontenc}
\usepackage[utf8]{inputenc}
\usepackage{microtype}
\usepackage{inconsolata}
\usepackage{graphicx}
\usepackage{xcolor}
\usepackage{enumitem}
\usepackage{booktabs}
\usepackage{xspace}

\usepackage{graphicx}

\usepackage{amsmath}
\usepackage{amssymb}
\usepackage[skip=0pt]{caption}
\usepackage{subcaption}
\usepackage{pifont}
\usepackage{booktabs}
\usepackage{algorithm}
\usepackage{algpseudocode}
\algrenewcommand\algorithmicrequire{\textbf{Input:}}
\algrenewcommand\algorithmicensure{\textbf{Output:}}
\usepackage{algpseudocode}
\usepackage{threeparttable}
\usepackage{multirow}
\usepackage{tabularx}
\usepackage{cleveref}
\usepackage{enumitem}
\usepackage{mathtools}
\usepackage{tcolorbox}
\usepackage{placeins}
\newtcolorbox{promptbox}{colback=gray!4,colframe=gray!50,boxrule=0.4pt,left=4pt,right=4pt,top=3pt,bottom=3pt,fontupper=\footnotesize}

\usepackage{tikz}
\usepackage{soul}
\usepackage{multirow}
\usepackage{colortbl}
\colorlet{lightgray}{White!30!lightgray}
\colorlet{lightblue}{White!70!MidnightBlue}

\usepackage{nicematrix}
\usepackage{arydshln}

\usepackage{listings}
\usepackage{caption}

\lstdefinestyle{papersafe}{
  captionpos=b,
  language=Python,
  basicstyle=\ttfamily\small,
  numbers=none,
  backgroundcolor=\color{code-bg},
  frame=single,
  rulecolor=\color{black},
  breaklines=True,
  breakatwhitespace=true,
  showstringspaces=false,
  keywordstyle=\bfseries\color{oi-blue},       
  commentstyle=\color{oi-green!60!black}, 
  stringstyle=\color{oi-red!85!black},         
  identifierstyle=\color{black},               
  emph={np,torch,nn,plt,fit,predict},          
  emphstyle=\color{oi-purple},
  ndkeywordstyle=\color{oi-orange},            
  aboveskip=2pt, belowskip=2pt, framesep=2pt, columns=fullflexible,
  prebreak=\mbox{\textcolor{gray}{$\hookleftarrow$}},
  postbreak=\mbox{\textcolor{gray}{$\hookrightarrow$}\space},
}
\crefname{figure}{Figure}{Figures}
\crefname{equation}{Equation}{Equations}
\crefname{appendix}{Appendix}{Appendices}
\crefname{table}{Table}{Tables}
\crefname{section}{Section}{Sections}

\newcounter{aq}
\newcommand{\aq}[1]{\refstepcounter{aq}\paragraph{Q\theaq: #1}}
\newcommand{\aqref}[1]{\S\ref{sec:analysis}, Q\ref{#1}}

\newcommand{\ours}{\textsc{ComplexIF}\xspace}

\title{\ours: A Challenging Multi-Turn, Multi-Constraint\\Instruction-Following Benchmark}

\author{Zhichao Xu\textsuperscript{1}\thanks{Equal contributions.} \quad
Bingsen Chen\textsuperscript{2}\footnotemark[1] \quad
Yawei Wang\textsuperscript{3}\footnotemark[1] \quad
Tao Li\textsuperscript{4} \quad \\
{\bf Haoyu Wang\textsuperscript{5} \quad 
Dan Roth\textsuperscript{5} \quad
Vivek Srikumar\textsuperscript{1}} \\
\\
\textsuperscript{1}University of Utah \quad
\textsuperscript{2}New York University \quad
\textsuperscript{3}George Washington University \quad \\
\textsuperscript{4}Google DeepMind \quad
\textsuperscript{5}University of Pennsylvania \\
}

\begin{document}
\maketitle
\begin{abstract}
As frontier LLMs saturate existing instruction-following benchmarks, measuring further progress requires harder tests that remain faithful to real use.
In practice, users often combine several constraints, add them at different turns of a conversation, and expect all of them to hold in the assistant's response.
We introduce \ours, a challenging benchmark for this setting, together with the automatic pipeline that generates it.
The pipeline expands seed requests into multi-turn conversations, weaves constraints into their turns, and keeps only tasks that a strong model can solve, so difficulty can be raised by adding constraints or turns while every task stays solvable.
\ours's two splits test complementary sides of instruction following: a verifiable split of format constraints, checked by programs, and a rubric split of semantic instructions, such as a tone or an audience, checked with an LLM judge.
Across 14 proprietary and open-weight models, even the strongest satisfies every constraint on fewer than half of the verifiable tasks, and no model leads on both splits.
Difficulty comes mainly from satisfying several constraints at once, reasoning helps without closing the gap, and rubric scores depend on the judge while model rankings stay stable.\footnote{\ours is released at \url{https://github.com/zhichaoxu-shufe/ComplexIF-Benchmark}.}
\end{abstract}

\section{Introduction}
\label{sec:intro}

Instruction following (IF), the ability to produce a response that satisfies user-specified constraints, is among the most fundamental capabilities of large language models, since it determines how well a model's outputs align with user intent~\citep{ouyang2022instructgpt,wei2022flan}.
In practice, users rarely ask for content alone: requests routinely carry constraints on the format, length, style, or wording of the output.
Measuring this capability reliably has therefore become a central concern, and the community has built an array of benchmarks for it.
One line of work uses verifiable constraints whose satisfaction a short Python program can decide, yielding objective and reproducible labels~\citep{zhou2023ifeval,pyatkin2025ifbench}.
A second line scores natural-language criteria with an LLM judge~\citep{zheng2023judging}, extending coverage to more implicit constraints, such as a tone, a persona, or a scope, that users routinely state but no program can reliably check~\citep{qin2024infobench,he2025advancedif}.
These benchmarks are now standard in model evaluation and serve as training signal for instruction tuning~\citep{viswanathan-2025-checklists-are-better,lambert2025tulu3}.
Each benchmark usually follows one of these two constraint types, evaluating either format constraints or semantic instructions.\footnote{We use the terms from AdvancedIF~\citep{he2025advancedif}, where semantic instructions are open-ended instructions, such as on tone or audience, that a program cannot easily verify.}

As models improve, however, these benchmarks are also saturating: frontier models now score near perfect on widely used benchmarks such as IFEval~\citep{grattafiori2024llama3,qwen2024qwen25,gemma2025gemma3}.
Current benchmarks also leave out three aspects of realistic use.
First, \emph{multi-turn context}: existing benchmarks are mostly single-turn, whereas real use is often a conversation, where constraints arrive at different turns and must keep holding as the conversation moves on.
Second, the \emph{number of constraints}: most tasks carry one or two constraints, whereas real requests often pack multiple constraints to be met jointly.
Third, \emph{both kinds of constraints}: real requests often mix format constraints with semantic instructions~\citep{he2025advancedif}, so a model should be tested on both.
We raise our central research question:

\vspace{3pt}
\begin{center}
\begin{minipage}{0.92\columnwidth}
\includegraphics[height=1em]{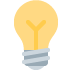}\itshape How can we build an instruction-following benchmark that is realistic and challenging, and that tests both format constraints and semantic instructions?
\end{minipage}
\end{center}
\vspace{3pt}

\begin{figure*}[t]
  \centering
  \includegraphics[width=\textwidth]{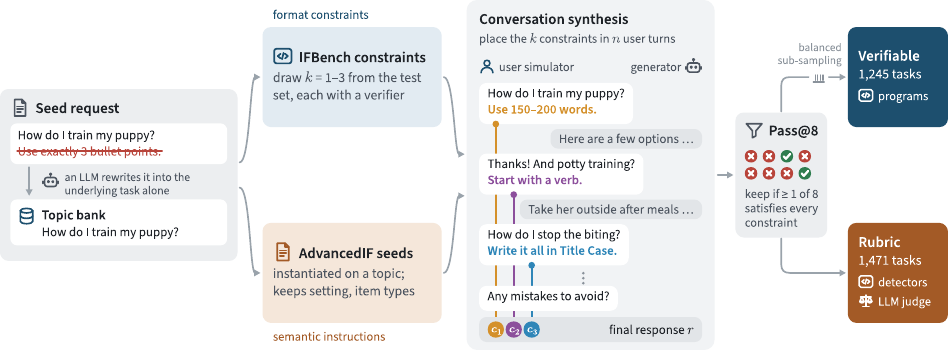}
  \caption{\textbf{\ours construction pipeline.} An LLM rewrites each seed request into the underlying task alone, building a topic bank shared by both splits (left). The verifiable split draws $k$ IFBench test constraints for a topic, and the rubric split instantiates an AdvancedIF seed on it (center left). Conversation synthesis places the constraints in $n$ user turns, which a user simulator and a generator play out; every constraint $c_j$ must hold in the final response $r$ (center right). Pass@8 keeps a task if one of 8 final responses satisfies every constraint (right).}
  \label{fig:pipeline}
\end{figure*}

To this end, we introduce \ours, a benchmark designed for multi-turn, multi-constraint instruction following.
The pipeline used to construct \ours expands seed requests from existing instruction-following datasets into multi-turn conversations, weaves constraints into different turns, and lets a strong model play out each conversation, keeping a task only if the model can satisfy every constraint.
The resulting tasks are \emph{realistic}, \emph{challenging}, \emph{reliable}, and \emph{scalable} (\S\ref{sec:pipeline}).
\ours comprises two splits that test complementary sides of instruction following: a verifiable split of explicit format constraints, each checked by a program, and a rubric split focusing on open-ended semantic instructions, checked with an LLM judge.

Across 14 proprietary and open-weight models, \ours proves challenging: even the strongest model satisfies every constraint on fewer than half of the verifiable tasks, and the majority of the models pass fewer than 30\% of tasks.
No model leads on both splits: Gemini 3.1 Pro is best on the verifiable split and Claude Opus 4.6 on the rubric split, so a strong instruction follower must do well on both.
Our analysis traces this difficulty to the number of constraints rather than to the length of the conversation.
Each additional constraint costs models a similar share of their performance: with three constraints, most models keep about a third of their single-constraint pass rate, whether that rate is 16\% or 41\%.
Conversation length, in contrast, has no detectable effect between four and ten turns, and constraints stated early are violated no more often than those stated recently; the difficulty lies in satisfying several constraints at once rather than in remembering them.
Reasoning helps substantially, raising Qwen3.5 27B from 22.3\% to 37.6\%, but does not close the gap.

Having both programmatic and LLM-judged labels also allows us to analyze how each aspect of instruction following should be measured.
LLM-judged scores are judge-dependent: replacing one judge with another shifts a model's score by 5--9 points, although the relative ranking of models stays stable.
We therefore recommend checking format constraints with programs, which give the deterministic verdict every time, and checking rubrics with an LLM judge while focusing on the relative ranking of models rather than over-interpreting the absolute scores.

To summarize, our contributions are:
\begin{itemize}[noitemsep,topsep=0pt,leftmargin=*]
  \item An automatic pipeline that turns seed requests into challenging multi-turn, multi-constraint instruction-following tasks, with independently controlled difficulty and every task solved by a strong model during construction (\S\ref{sec:pipeline}).
  \item \ours, a benchmark with complementary splits for format constraints and semantic instructions, together with an evaluation of 14 proprietary and open-weight models showing that it remains far from solved (\S\ref{sec:benchmark}, \S\ref{sec:experiments}).
  \item An in-depth analysis of what makes multi-turn, multi-constraint instruction following hard, including the number of constraints, conversation length, and reasoning, and of how the choice of judge affects measured performance (\S\ref{sec:analysis}).
\end{itemize}

\section{Related Work}
\label{sec:related}
\paragraph{Format constraints.}
A widely used way to measure instruction following attaches format constraints whose satisfaction a short program can decide, such as a word count, a required keyword, or an output layout.
IFEval~\citep{zhou2023ifeval} introduced 25 such constraint types; IFBench~\citep{pyatkin2025ifbench} added 58 new ones and kept separate training and test constraints to measure generalization; and COLLIE~\citep{yao2024collie} composes constraints at the word, sentence, and paragraph level.
IFEval++~\citep{dong2026revisiting} adds prompt perturbations to IFEval and shows that existing LLMs degrade to different degrees.
These benchmarks are mostly single-turn, and most of their prompts carry one or two constraints.
The verifiable split of \ours follows this style: it draws its constraints from the IFBench test set and combines several of them across the turns of a conversation.

\paragraph{Semantic instructions and rubric-based evaluation.}
Many instructions users give are open-ended, such as a tone to keep, a character to play, or an audience to write for. These constraints can be challenging to verify deterministically via Python programs. 
AdvancedIF~\citep{he2025advancedif} describes instruction following as ranging from simple format constraints to such semantic instructions, and evaluates it with rubrics: it pairs human-written prompts, including multi-turn and system-prompt settings, with human-written yes/no criteria that an LLM judge checks, and also trains models with rubric-based rewards.
Earlier benchmarks also break requests into requirements that an LLM judge checks: FollowBench~\citep{jiang2024followbench} adds constraints to an instruction one level at a time, ComplexBench~\citep{wen2024complexbench} composes them through structures such as chains and selections, and InFoBench~\citep{qin2024infobench} breaks each instruction into yes/no requirements.
Such rubric-based evaluation usually relies on an LLM judge to check each item~\citep{ribeiro-etal-2020-beyond,arora2025healthbenchevaluatinglargelanguage,chiang2024chatbotarena,liu2026surveyofrubrics}, so its scores can carry that judge's biases.
We build the rubric split of \ours based on AdvancedIF seeds, and we also measure how much its scores depend on the judge (\aqref{sec:analysis-measurement}).

\paragraph{Multi-turn and system-prompt instruction following.}
Several benchmarks evaluate instruction following over a conversation.
Multi-IF~\citep{he2024multiif} extends IFEval prompts to three turns in several languages, MT-Bench-101~\citep{bai2024mtbench101} scores a taxonomy of multi-turn abilities, and StructFlowBench~\citep{li2025structflowbench} labels the structural relations between turns, such as follow-ups and refinements.
MultiChallenge~\citep{deshpande2025multichallenge} targets failures such as forgetting an instruction from an earlier turn, and grades each conversation with an instance-level rubric.
Two recent benchmarks extend conversations further and score every turn: EvolIF~\citep{jia2026evolif} keeps adding and changing topics and constraints, and ends a conversation once the model fails several turns in a row; SEQUOR~\citep{canaverde2026sequor} builds 50-turn conversations from constraints found in real user chats and finds that accuracy falls as the conversation grows.
A related line studies instructions given in the system prompt: SysBench~\citep{qin2025sysbench} tests whether models keep following system messages across turns, and IHEval~\citep{zhang2025iheval} tests whether they respect the priority of system instructions over user instructions when the two conflict.
\ours brings these settings together: each task carries several constraints stated at different turns (and, in the rubric split, in the system prompt), and only the final response is scored, so every constraint must still hold at the end of the conversation.

\paragraph{LLM-as-a-judge.}
LLM judges are widely used to score open-ended responses~\citep{zheng2023judging} because they are easy to use, and prior work has discussed their limitations. For example, \citet{wang2024fair,zheng2023judging} show that judges prefer the first of two responses and longer responses, and they may prefer their own generations~\citep{panickssery2024selfpref}.
Rubric benchmarks narrow the judge's task to yes/no questions about one criterion at a time~\citep{qin2024infobench,deshpande2025multichallenge}, yet each reported score still comes from a single judge.
We score identical responses with three judges: two proprietary models from different providers and one open-weight model, and find that the choice of judge shifts absolute scores by 5--9 points while leaving the ranking of models nearly unchanged (\aqref{sec:analysis-measurement}).

\section{Benchmark Construction}
\label{sec:construction}

\subsection{Task Formulation}
\label{sec:formulation}

\paragraph{Task.}
A task in \ours is a conversation $x = (s, u_1, a_1, \dots, u_{n-1}, a_{n-1}, u_n)$ with an optional system prompt $s$, $n$ user turns $u_i$, and the $n-1$ assistant turns $a_i$ that precede the last user turn.
The task carries a set of $k$ constraints $C = \{c_1, \dots, c_k\}$.
Constraint $c_j$ is stated at turn $t_j \in \{0, 1, \dots, n\}$, where $t_j = 0$ denotes the system prompt, and stays in force for the rest of the conversation.
Given $x$, the evaluated model writes the final response $r$; the earlier assistant turns are fixed context, written when the benchmark was built (\S\ref{sec:pipeline}).
Each constraint has a verifier $v_j(x, r) \in \{0, 1\}$, and the task is passed only if every verifier accepts $r$:
\begin{equation}
  \mathrm{pass}(x, r) = \prod_{j=1}^{k} v_j(x, r).
\end{equation}
Our main metric is the task-level pass rate, the share of tasks passed.
We also report the constraint-level pass rate, the share of all constraints satisfied, showing how often an individual constraint is met.
The number of constraints $k$ and the number of user turns $n$ are the two difficulty dimensions of focus.

\paragraph{Constraint types.}
Constraints differ in how $v_j$ is computed.
A verifiable constraint, such as ``include each of these keywords a given number of times,'' is paired with a program that decides $v_j$ from $r$, so its label is exact and reproducible.
A rubric item, such as ``keep the persona of a tour guide,'' is a natural-language criterion; it is decided by a deterministic detector when one applies and by an LLM judge otherwise.
Rubric items cover more implicit requirements, at the cost of depending on the judge.
The two kinds of constraints define the two splits of \ours (\S\ref{sec:benchmark}).

\subsection{Generation Pipeline}
\label{sec:pipeline}

\paragraph{Design goals.}
We design the pipeline (Figure~\ref{fig:pipeline}) around the four properties of \S\ref{sec:intro}.
Tasks are \emph{realistic} because every topic comes from an existing user request and every constraint is rewritten into the language of a user turn.
They are \emph{challenging} because several constraints are stated across the conversation and must all hold in the final response.
They are \emph{reliable} because every constraint keeps its verifier, rewrites must preserve the exact values the verifier checks, and the pass@8 filter keeps only tasks a strong model can solve.
And they are \emph{scalable} because the number of constraints and the number of turns are set independently.

\paragraph{Topic bank construction.}
Each task of \ours starts from a seed request sourced from an existing instruction-following dataset.
These requests often carry their own output constraints, so we prompt an LLM to rewrite each one into the underlying task alone, with every output requirement removed.
The result is a bank of topics that users have asked about, and none of their original constraints carry over into \ours.
\paragraph{Constraint sampling.}

For a target number of constraints $k$, we draw $k$ constraints from a constraint bank where every constraint comes with a verifier, and sample the parameters each one needs, such as keywords, counts, or a target phrase.
Some constraints require the response to consist only of their own output, and these cannot be combined with any other constraint.
For example, ``Answer with one of the following options: a), b), c), d). Do not give any explanation.'' leaves no room for a constraint such as a minimum word count.
We manually mark six such constraints and never draw them together with another constraint.
\paragraph{Conversation synthesis.}
A user-simulator LLM first plans the user side of an $n$-turn conversation from the topic, so that the task develops from turn to turn as a user's request would.
The simulator asks only for content and never specifies an output format, so the user turns cannot contradict a sampled constraint. It then adapts its planned user turn based on the assistant's responses, making sure the conversation flow is natural and not mechanical.
\paragraph{Constraint placement.}

Each constraint is assigned to one of the planned user turns, and an LLM rewrites the constraint into that turn so that it reads naturally.
We then check the rewritten turn with a program, and if the rewrite changes or drops any number or quoted phrase of the original constraint, we discard the task, so every kept turn states exactly what the verifier checks.
\paragraph{Pass@8 solvability filter.}
A strong generator model then plays out the conversation with the simulator.
The generator writes each assistant turn conditioned on the conversation so far, and before each later user turn is sent, the simulator rewrites the planned turn to respond to what the assistant actually said, for example by pointing out that an answer missed the request.
The rewrite must keep the planned request and every number and quoted phrase of its constraints; if any of them is lost, we keep the planned turn as it was.
In practice, the sampled constraints can conflict with the topic or be too hard for the generator to solve.
We therefore draw up to eight final responses to the played-out conversation and keep the task only if at least one of them satisfies every constraint.
Tasks with no satisfying response are discarded. This way, we ensure every task in~\ours can be solved by a frontier model at the time of construction (Appendix~\ref{sec:appendix-checkers} notes three tasks affected by a later checker update).
\paragraph{Controlling difficulty.}
Since the number of constraints $k$ and the number of user turns $n$ are set independently, we can create new tasks by raising either one without new annotation.
\subsection{The \ours Benchmark}
\label{sec:benchmark}
We construct \ours with two complementary splits: a verifiable split of format constraints that a program checks, and a rubric split of semantic instructions checked against rubrics.
Table~\ref{tab:stats} shows details of the two splits.
\begin{table}[t]
  \centering
  \setlength{\tabcolsep}{4pt}
  \resizebox{\columnwidth}{!}{%
  \begin{tabular}{@{}lcc@{}}
    \toprule
    & Verifiable & Rubric \\
    \midrule
    Tasks & 1,245 & 1,471 \\
    Constraints / items & 2,132 & 10,003 \\
    \quad per task & 1--3 & 1--20 (median 6) \\
    User turns per task & 4--10 & 1--10 \\
    Tasks with system prompt & -- & 529 \\
    Seed source & \begin{tabular}[c]{@{}c@{}}InFoBench,\\Multi-IF,\\StructFlowBench\end{tabular} & AdvancedIF \\
    Generator & \begin{tabular}[c]{@{}c@{}}GPT-5.4 (1,010),\\Opus 4.6 (235)\end{tabular} & Opus 4.6 \\
    Label & program & \begin{tabular}[c]{@{}c@{}}detector (2,044),\\LLM judge (7,959)\end{tabular} \\
    \bottomrule
  \end{tabular}%
  }
  \caption{Statistics of the \ours splits. Rubric-split label counts give how many items are checked by a deterministic verifier versus the LLM judge.}
  \label{tab:stats}
\end{table}

\paragraph{Verifiable split.}
We source seed requests from InFoBench~\citep{qin2024infobench}, StructFlowBench~\citep{li2025structflowbench}, and the first turns of Multi-IF~\citep{he2024multiif}.
We source the constraints from the IFBench test set~\citep{pyatkin2025ifbench}; each constraint can be verified deterministically by a Python program.
Each task carries 1 to 3 constraints and 4 to 10 user turns.
The first constraint is always stated in the opening request, and the others are stated at later user turns.
We use two strong proprietary LLMs, GPT-5.4 and Claude Opus 4.6, as the generators.
We use 54 out of 58 constraints from IFBench, dropping four constraints as they cannot be combined with freeform topics or other constraints.
After the pass@8 solvability filter, the resultant tasks are unbalanced: a few constraint types dominate, most tasks carry a single constraint, and some conversation lengths (number of turns) are more frequent than others.
We therefore re-balance the split by subsampling.
This yields 1,245 tasks with a more even spread of constraint types, numbers of constraints, and conversation lengths (Figure~\ref{fig:subsampling}; detailed pipeline in Appendix~\ref{sec:appendix-subsampling}).
\paragraph{Rubric split.}
We start from 600 seed prompts from AdvancedIF~\citep{he2025advancedif}, each with a list of human-written rubric items.
We use Claude Sonnet 4.5 to label each seed rubric item as either verifiable by a program or requiring an LLM judge.
We then instantiate each seed on new topics from the same topic bank and keep its setting: a single complex request, a multi-turn conversation that carries constraints across turns, or a system prompt that states standing instructions.
For each new task, Claude Opus 4.6 writes the user turns and a new list of binary rubric items, each of the same type as the corresponding seed item.
Items labeled verifiable are first matched against deterministic detectors, such as checks for format markers, word counts, or required phrases; the remaining items go to an LLM judge, which grades all of a task's judged items in one call.
We use Claude Opus 4.6 as the generator. Since the user turns are written together with the task, each of up to eight draws regenerates every assistant turn, and we keep the task only if at least one response satisfies all rubric items.
The final rubric split contains 1,471 tasks built from 517 of the seed prompts.

\begin{figure*}[t]
  \centering
  \includegraphics[width=\textwidth]{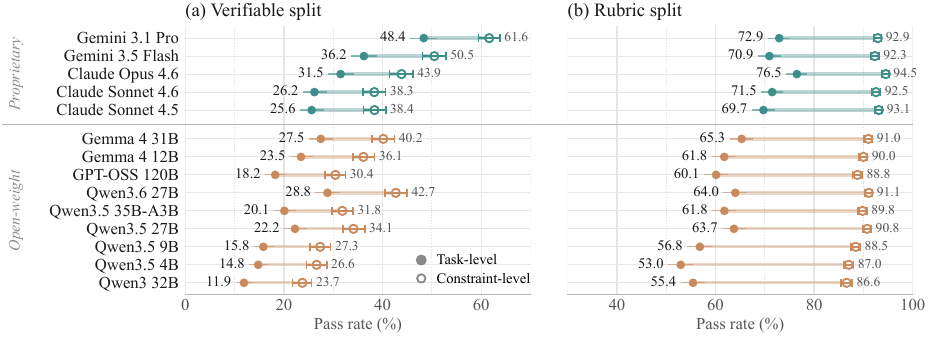}
  \caption{\textbf{Main Results.} We report the task-level (filled dots) and constraint-level (hollow dots) pass rates on the (a)~verifiable and (b)~rubric splits. Error bars indicate 95\% confidence intervals.}
  \label{fig:main_results}
\end{figure*}

\section{Experiments}
\label{sec:experiments}

\subsection{Experimental Setup}
\label{sec:setup}
\paragraph{Models.}
We evaluate 14 models of varying sizes, versions, and model families. For proprietary models, we include Claude Opus 4.6, Sonnet 4.5 \& 4.6, Gemini 3.1 Pro and 3.5 Flash. The other nine are open-weight: Gemma 4 (12B and 31B), Qwen3 32B, Qwen3.5 (4B, 9B, 27B, and 35B-A3B), Qwen3.6 27B, and GPT-OSS 120B.

\paragraph{Evaluation Setup.}
Each model receives the conversation up to the final user turn, including the earlier assistant turns, and generates the final response considering all previous turns. We run open-weight models at temperature 0 for reproducibility, with thinking off; GPT-OSS 120B cannot turn reasoning off, so it runs at medium reasoning effort. Proprietary models run with thinking on, at the temperatures given in Appendix~\ref{sec:appendix-settings}. We study the effect of reasoning separately in \aqref{sec:analysis-reasoning}.
We report the task-level and constraint-level pass rates defined in \S\ref{sec:formulation} on both splits.
On the verifiable split, a program checks every constraint.
On the rubric split, each rubric item counts as one constraint, and we adopt an LLM judge to verify the rubric items, averaging the scores of three judges for robustness: Claude Opus 4.6, GPT-5.5, and GPT-OSS 120B.
Appendix~\ref{sec:appendix-settings} gives the full decoding settings.

\subsection{Main Results}
\label{sec:results}

Figure~\ref{fig:main_results} reports task-level and constraint-level pass rates for all models on both splits. We observe that:

\noindent\textbf{Multi-turn, multi-constraint instruction following remains far from solved, even for frontier models.}
The strongest model, Gemini 3.1 Pro, satisfies every constraint on only 48.4\% of verifiable tasks, and the average over all 14 models is 25.1\%.
Frontier models now approach the ceiling of single-turn benchmarks such as IFEval~\citep{grattafiori2024llama3,qwen2024qwen25,gemma2025gemma3}, yet this ability does not carry over to the multi-turn, multi-constraint setting.
Neither scale nor newer versions close the gap: Qwen3.5 rises only from 14.8\% at 4B to 22.2\% at 27B, Claude Sonnet 4.6 barely improves on Sonnet 4.5, and the best open-weight model, Qwen3.6 27B, already matches both Sonnet models.
\ours thus remains challenging across model families, sizes, and versions.

\noindent\textbf{Models struggle not only to combine constraints but to satisfy individual ones.}
We observe that the constraint-level pass rate is also low: models satisfy 37.5\% of individual constraints on average, and only the two Gemini models exceed half.
These constraints demand precise control over the surface form of a response, such as placing a keyword at an exact word position or keeping the length within a given range, and that control must hold while the response still serves the user's request.
Precise control over form is thus a first source of difficulty, present before any constraints are combined.

\noindent\textbf{Models satisfy semantic instructions far more often than format constraints, yet neither split is close to solved, and no model leads on both.}
On the rubric split, models satisfy 90.6\% of individual rubric items, more than twice the rate for verifiable constraints, since rubric items describe the kind of requirements users routinely state, such as keeping a persona or staying within a scope.
Yet rubric tasks remain far from solved: a task carries 6.8 items on average, so even rare misses compound, and task-level pass rates range from only 53.0\% to 76.5\%.
We also note model rankings on the two splits are different: Gemini 3.1 Pro leads the verifiable split and Claude Opus 4.6 the rubric split, and Claude Sonnet 4.6 ranks sixth on the verifiable split but third on the rubric split, although the two rankings agree overall (Spearman $\rho = 0.92$).
We believe that the two splits test complementary sides of instruction following, and doing well on one does not automatically imply doing well on the other (see Appendix~\ref{sec:appendix-cases} case studies).

\noindent\textbf{Model rankings are consistent with existing single-turn benchmarks, while task-level pass rates are much lower.}
For the seven open-weight models that we also run on IFBench~\citep{pyatkin2025ifbench} (Table~\ref{tab:ifbench}), the ranking on \ours largely agrees with the IFBench ranking (Spearman $\rho = 0.82$ under the comparable prompt-level strict accuracy), yet every model scores only about half its IFBench accuracy.
\ours therefore measures the same underlying capability as existing verifiable benchmarks, but leaves substantial headroom for measuring progress.

\begin{figure*}[t]
  \centering
  \includegraphics[width=\textwidth]{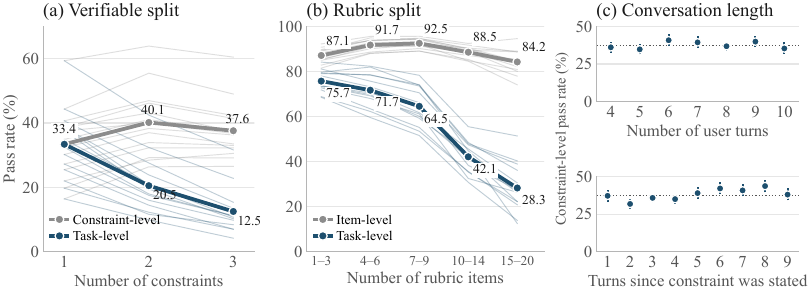}
  \caption{\textbf{Pass Rate by Number of Constraints and Conversation Length.} (a)--(b)~Task-level and constraint-level pass rates by number of constraints on the verifiable split and by number of rubric items on the rubric split; thin lines are individual models, thick lines their average. (c)~Constraint-level pass rate on the verifiable split by number of user turns (top) and turns since the constraint was stated (bottom), averaged over the 14 models and adjusted for constraint type; error bars are 95\% bootstrap confidence intervals over tasks, and the dotted line is the overall mean.}
  \label{fig:num_constraints}
  \label{fig:conv_length}
  \vspace{-10pt}
\end{figure*}

\subsection{Analysis}
\label{sec:analysis}
\aq{How does performance scale with the number of constraints?}\label{sec:analysis-constraints}
We vary the number of constraints on the verifiable split and group rubric tasks by number of rubric items (Figure~\ref{fig:num_constraints}a--b; Tables~\ref{tab:num_constraints},~\ref{tab:num_constraints_rubric}). We find that:
\textbf{task-level pass rate drops sharply with each added constraint, at a similar relative rate across models and on both splits.}
On the verifiable split, the average task-level pass rate falls from 33.4\% with one constraint to 20.5\% with two and 12.5\% with three; on the rubric split, it falls from 75.7\% for tasks with one to three items to 28.3\% for tasks with 15 to 20.
Strikingly, the relative cost of an added constraint barely depends on model strength: although single-constraint pass rates range from 16.4\% to 40.7\% across the Claude and open-weight models, each of them keeps about a third of that rate at three constraints, and only the two Gemini models degrade more gracefully, keeping about half.
Stronger models thus start from a higher level but are, with few exceptions, no more robust to accumulating constraints.

\noindent\textbf{The performance decline comes from satisfying multiple constraints jointly, as the constraint-level pass rate stays flat.}
Adding constraints does not make each one harder to satisfy: the constraint-level pass rate on the verifiable split stays between 33.4\% and 40.1\% as constraints are added, and the share of rubric items satisfied stays between 84\% and 93\% across task sizes.
What grows is the chance that at least one constraint is missed, so a task's pass rate shrinks roughly multiplicatively with its number of constraints.
Satisfying individual constraints is hard; satisfying several jointly is harder, and this compounding is what drives the difficulty of \ours.

\aq{Does more reasoning help?}\label{sec:analysis-reasoning}
\begin{figure}[t]
  \centering
  \includegraphics[width=\columnwidth]{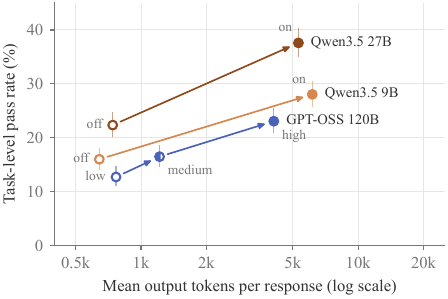}
  \caption{\textbf{Effect of Reasoning.} We report the task-level pass rate on the verifiable split against the mean number of output tokens (log scale), with arrows going from each model's lowest to highest reasoning setting. Error bars indicate 95\% confidence intervals.}
  \label{fig:reasoning_tokens}
\end{figure}

We compare thinking off and on for Qwen3.5 9B and 27B, as well as low, medium, and high reasoning effort for GPT-OSS 120B, on the verifiable split (Figure~\ref{fig:reasoning_tokens}; Appendix~\ref{sec:appendix-settings-reasoning}). We find that: \textbf{more reasoning consistently improves instruction following, under both binary and graded reasoning controls.}
Turning thinking on raises Qwen3.5 9B from 16.0\% to 28.0\% and Qwen3.5 27B from 22.3\% to 37.6\%, and GPT-OSS 120B improves steadily from 12.7\% at low effort to 16.5\% at medium and 23.1\% at high.
These gains far exceed those from scale: thinking adds 12 to 15 points to Qwen3.5, whereas growing the model from 4B to 27B adds about 7.
A plausible explanation is that reasoning gives the model room to plan its response around the constraints and to check them before answering, which is exactly what joint satisfaction (Q\ref{sec:analysis-constraints}) requires.

These gains come at a significant token usage.
Thinking raises the mean response length of Qwen3.5 from about 700 tokens to 5,300--6,100, and GPT-OSS 120B uses 4,088 tokens per response at high effort versus 767 at low, yet even the best setting passes fewer than 40\% of tasks.

\aq{Do models lose track of constraints as conversations grow?}\label{sec:analysis-length}
We measure how often constraints are satisfied as a function of conversation length and of how many turns before the final response a constraint was stated, for conversations of four to ten user turns (Figure~\ref{fig:conv_length}c). We find that \textbf{within four to ten turns, models keep their constraint satisfaction rate as the conversation grows and however early a constraint was stated.}
Across conversations of four to ten user turns, the constraint-level pass rate shows no trend, and constraints stated in the opening turn are satisfied as often as those stated just before the final response.
Models therefore keep track of constraints throughout the conversation; what they fail at is satisfying all of them at once (Q\ref{sec:analysis-constraints}).
On \ours, the challenge of multi-turn instruction following lies in the accumulation of requirements rather than in remembering them.
We note much longer conversations may behave differently: SEQUOR~\citep{canaverde2026sequor} finds that accuracy falls over 50-turn conversations, most of all when constraints are added over time.

\aq{How much do rubric scores depend on the judge?}\label{sec:analysis-measurement}
\begin{figure}[t]
  \centering
  \includegraphics[width=\columnwidth]{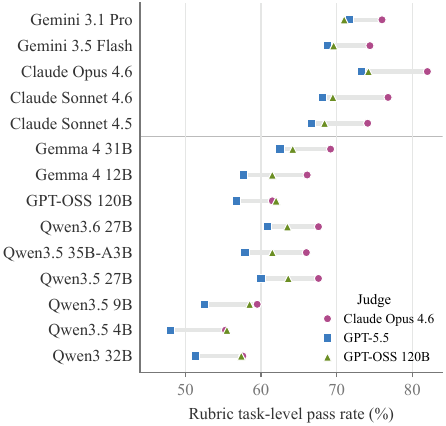}
  \caption{\textbf{Judge Model Comparison.} We report the task-level pass rate on the rubric split when three LLM judges score the same responses.}
  \label{fig:judge_scores}
  \vspace{0pt}
\end{figure}

We score the same rubric-split responses with three distinct judges, Claude Opus 4.6, GPT-5.5, and GPT-OSS 120B, and compare their verdicts (Figure~\ref{fig:judge_scores}; Appendix~\ref{sec:appendix-judges}). We find that: \textbf{the choice of judge significantly shifts rubric scores, yet leaves the ranking of models nearly unchanged.}
Depending on the judge, a model's rubric score shifts by 7.1 points on average and by up to 8.7, and the direction is consistent: Claude Opus 4.6 is the most lenient judge for 12 of the 14 models, and GPT-5.5 the strictest for 13.
Yet all three judges rank the models almost identically, with pairwise Spearman correlations of at least 0.97.
Rubric scores are therefore meaningful for comparing models under a fixed judge, but an absolute rubric score cannot be interpreted without knowing which judge produced it.

\noindent\textbf{Judges frequently disagree on whether an individual rubric item is met, even though their aggregate scores rank models alike.}
Two judges reach the same verdict on 94--96\% of items, but since about 90\% of items pass under any judge, much of this agreement would occur by chance; corrected for chance, agreement is only moderate (Cohen's $\kappa$ of 0.53--0.64).
These disagreements likely average out over thousands of items, which would explain why rankings stay stable, but individual verdicts remain noisy.
We therefore report rubric scores averaged over three judges and compare models only under the same judges, while format constraints are checked exactly by programs.

\section{Conclusion}
\label{sec:conclusion}
We introduced \ours, a benchmark for multi-turn, multi-constraint instruction following. 
Its two splits test complementary sides of instruction following: format constraints, checked by programs, and semantic instructions, checked against rubrics.
Across 14 proprietary and open-weight models we evaluated, neither split is close to solved, even for frontier models.
The difficulty of \ours comes mainly from satisfying several constraints at once: within four to ten turns, models keep track of constraints as the conversation grows, yet each added constraint lowers their task-level pass rate.
Evaluations on \ours reveal that multi-turn, multi-constraint instruction following remains challenging for current LLMs, leaving substantial headroom for future models.

\section*{Limitations}
We evaluate 14 proprietary and open-weight models that were available when we ran our experiments, including Claude Opus 4.6 and Gemini 3.1 Pro. New models appear every few weeks and we cannot evaluate all of them, but scoring a new model needs only its final responses and the same checkers and judges, so its results are directly comparable to ours. The verifiable split covers one to three constraints and four to ten user turns; since the pipeline sets both independently and needs no new annotation, we can leverage our pipeline to generate harder versions as models improve. Our constraints are sourced from IFBench, our rubric seeds from AdvancedIF, and our rubric scores from three judges, and other constraint sets, rubric sources, and judges fit into the same pipeline. Finally, we study reasoning on three open-weight models whose reasoning setting we can control, and the same protocol applies to any model that exposes such a setting.

\section*{Ethical Considerations}
This paper presents a benchmark for evaluation LLM's instruction following behaviors. All artifacts used in this paper have been disclosed in Appendix~\ref{sec:appendix-licenses} about licenses. To the best of our knowledge, this paper does not raise ethical risks.

\bibliography{custom}

\newpage
\appendix
\section{Construction Details}
\label{sec:appendix-construction}
\subsection{Prompts}
\label{sec:appendix-prompts}

We use five LLM prompts to build \ours, and we give all five in Figures~\ref{fig:prompt-topic}--\ref{fig:prompt-rubric}.
\begin{figure}[!htb]
  \centering
  \includegraphics[width=\columnwidth]{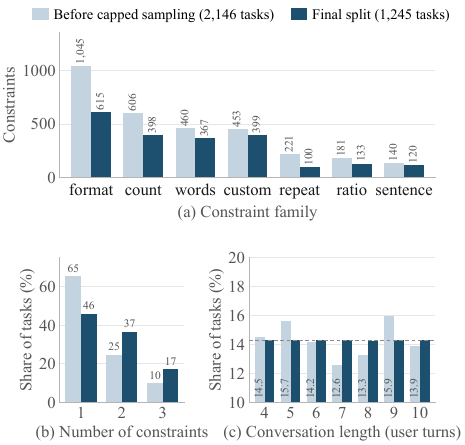}
  \caption{\textbf{Balanced Sub-sampling.} We report (a)~the number of constraints per constraint family and the share of tasks by (b)~number of constraints and (c)~conversation length, before and after capped sampling. The dashed line marks a roughly equal share across 7 lengths (14.3\%), which every conversation length reaches in the final split.}
  \label{fig:subsampling}
\end{figure}

\paragraph{Topic extraction.}
We strip the output constraints from each seed request with one prompt that works across all three source datasets, so we never write dataset-specific parsing code.
We drop a seed if the model returns an empty task, which happens when the original prompt was just a constraint without actual ``topic''.
Figure~\ref{fig:prompt-topic} gives the prompt.

\begin{figure*}[t]
\begin{promptbox}
\textbf{System.} You extract the bare \emph{carrier task} from an instruction-following prompt. A carrier task is the underlying request a user wants done (e.g.\ `write a product description', `summarize the passage', `plan a 3-day trip') with ALL output constraints stripped away --- remove any requirement about length, format, wording, structure, keywords, tone, casing, sections, JSON, bullet/number counts, forbidden words, or similar. Keep only the substance of what to produce and any input content it operates on. You also give a short domain label (1--3 words).

\smallskip
\textbf{User.} Read this instruction-following prompt and extract its carrier task with all output constraints removed.
PROMPT: \{prompt\}
Return ONLY a JSON object, no prose: \{"task": "\ldots", "domain": "\ldots"\}.
Rules: Strip every output-format / length / wording / structure constraint. Preserve the core request and any input material it must act on. If the prompt is purely a constraint with no real task, set task to "". \texttt{domain} is a topical label like ``cooking'', ``software'', ``travel'', ``finance''.
\end{promptbox}
\caption{Prompt for topic extraction.}\label{fig:prompt-topic}
\end{figure*}

\paragraph{User simulator.}
We plan the user turns up front from the topic alone, which keeps the topic coherent, and then place the constraints into the planned turns (\S\ref{sec:pipeline}).
The planning prompt asks only for content, so no planned turn asks for an output format that could contradict a constraint.
We then adapt each planned turn while the conversation is played out: before a turn is sent, the simulator sees the conversation so far and the planned turn, and rewrites the turn so that it responds to what the assistant actually said.
We skip the first turn, since there is no reply to respond to yet.
We keep the planned turn unchanged if the rewrite drops any number or quoted phrase of a constraint stated in that turn, or if it is cut off or written as the assistant instead of the user.
We use the simulator only for the verifiable split; in the rubric split, the user turns are written together with the rubric (see \emph{Rubric instantiation} below).
Figure~\ref{fig:prompt-simulator} gives the planning prompt (top) and the adaptation prompt (bottom).

\begin{figure*}[t]
\begin{promptbox}
\textbf{User.} Write the USER side of a natural \{n\}-turn conversation in which a user works with an AI assistant on an evolving task. Topic: ``\{topic\}''. Each user turn should build on the conversation so far --- asking a follow-up, adding a requirement about the SUBJECT, or refining what is being asked --- as a real user would across a working session.
CONTENT ONLY --- do NOT specify output form. Ask for information, analysis, drafts, revisions or opinions. Never state how the answer should be shaped: no tables, bullet points, numbered lists, outlines, sections, headings, JSON, CSV, markdown, code blocks, word or sentence counts, paragraph counts, tone or style directives. Formatting requirements are supplied separately and a turn that prescribes form can contradict them. ``Explain the causes'' is good; ``explain the causes in a 3-column table'' is not.
FORMAT (no prose before or after, no assistant lines): User: $\langle$turn 1 --- opening request about the topic$\rangle$ User: $\langle$turn 2 --- a natural follow-up that builds on turn 1$\rangle$ \ldots\ (\{n\} ``User:'' lines total)
Write exactly \{n\} ``User:'' lines and NOTHING else. Do not write any ``Assistant:'' lines.
\end{promptbox}
\smallskip

\begin{promptbox}
\textbf{System.} You are writing the USER side of a conversation with an AI assistant, one turn at a time.
You will be given the conversation so far and a DRAFT of the user's next message. The draft was written in advance, before the assistant's replies existed, so it may assume things the assistant never actually said.
Your job is to rewrite the draft so it reads like a real person responding to what the assistant ACTUALLY said, while asking for the same thing the draft asks for.

RULES:
1.~Keep the draft's request. If the draft asks for a summary, ask for a summary. Do not change the topic or substitute a different task.
2.~Preserve EVERY requirement, rule, constraint, number, quoted phrase and single-character literal from the draft EXACTLY as written, character for character. These are the specification the assistant is being held to; changing ``at least 3'' to ``at least three'', or dropping a quoted phrase, corrupts it.
3.~Fix any false claim about what happened. If the draft thanks the assistant for something it never provided, or says an answer was helpful when it did not answer, replace that with what a real person would say instead --- pointing out the problem, asking again, or simply moving on without comment.
4.~Do not praise or thank the assistant for work it did not do.
5.~Stay in the user's voice. You are the person asking, never the assistant, and never a narrator.
6.~Keep it about the same length as the draft. Do not add new requirements that were not in the draft.

Output ONLY the rewritten user message. No preamble, no explanation, no quotation marks around the whole thing, no ``User:'' prefix.

\smallskip
\textbf{User.} CONVERSATION SO FAR: \{conversation\}
DRAFT OF THE USER'S NEXT MESSAGE: \{planned turn\}
Rewrite the draft so it responds to what the assistant actually said, keeping the same request and preserving every requirement verbatim. Output only the rewritten message.
\end{promptbox}
\caption{User-simulator prompts: planning the user turns (top) and adapting each planned turn to the assistant's reply (bottom).}\label{fig:prompt-simulator}
\end{figure*}

\paragraph{Constraint rewrite.}
We ask a model to weave each constraint into the user turn that introduces it, as natural request language in the style of IFEval prompts.
We then check the rewrite with a program and drop the task if any number or quoted phrase from the original constraint is missing or changed.
Figure~\ref{fig:prompt-rewrite} gives the prompt.

\begin{figure*}[t]
\begin{promptbox}
\textbf{System.} You rephrase instruction-following prompts so that explicit constraint bullet points are woven into natural prose instead of being listed in a separate ``Rules I must follow:'' block.

HARD CONSTRAINTS:
1.~Preserve EVERY constraint in meaning AND in exact surface form of any numeric value, keyword, literal phrase, pattern, or code-like token. Examples that MUST appear verbatim in the rewrite: numeric values (``exactly 3'', ``at most 100 words'', ``at least 2 sections''); listed keywords / forbidden words (['hello', 'world'] must stay as those exact words); literal phrases (``I know or I don't know'', ``yes/no/maybe''); ratios, positions, increments phrased numerically.
2.~NEVER convert a number between digit and word form. Every number that appears as digits in a constraint MUST appear as those SAME digits in your rewrite --- keep ``8'' as ``8'' (never ``eight''), ``27'' as ``27'' (never ``twenty-seven''), ``1'' as ``1'' (never ``one'', ``once'', ``a'', or ``a single''). This holds even when spelling it out or omitting it would read more naturally: fluency does not override this. Do not convert word-form numbers to digits either.
3.~Reproduce every quoted or single-character literal EXACTLY, including its case and any surrounding punctuation --- e.g.\ 'A' stays the character A, 'Z' stays Z; do not paraphrase ``from 'A' to 'Z{'}'' into ``the alphabet''.
4.~Do NOT drop any constraint.
5.~Do NOT add new constraints that weren't in the input.
6.~Preserve the original underlying user task (the non-rules part of the message) --- do not rewrite it into a different task.
7.~Keep a single user message. No bullet list. No ``Rules I must follow:'' header.

Before finishing, silently re-check that every digit, quoted string, and single-character literal from the constraints appears verbatim in your rewrite. If any is missing, fix it.
Output format: return ONLY the rewritten user message. No preamble, no explanation, no code block fences.

\smallskip
\textbf{User.} Rewrite the following prompt in IFEval prose style: one natural paragraph (or a short instruction followed by a constraint-dense sentence) where every constraint is woven into the fluent instruction text. Make it sound like a real user asking for help --- no bullet list of rules.
ORIGINAL TASK (preserve this task exactly): \{task\}
CONSTRAINTS TO WEAVE IN (each must appear, preserving all numeric values / keywords / literal phrases verbatim --- keep every number in DIGIT form, e.g.\ ``8'' not ``eight'', ``1'' not ``one''/``a single''; keep single-character literals like 'A' / 'Z' exactly): \{constraints\}
Return ONLY the rewritten user message.
\end{promptbox}
\caption{Prompt for writing constraints into a user turn.}\label{fig:prompt-rewrite}
\end{figure*}

\paragraph{Rubric instantiation.}
For the rubric split, we give the model one AdvancedIF seed, its conversation and its rubric, and ask for a new task on a new topic with the same shape.
Before this step, we use Claude Sonnet 4.5 to label every seed rubric item as program-checkable or judged.
We pass in the number of user turns, whether a system prompt is needed, and the label of each seed item, so the new rubric keeps the seed's mix of program-checkable and judged items.
The topic comes from the same topic bank as the verifiable split.
Figure~\ref{fig:prompt-rubric} gives the prompt.

\begin{figure*}[t]
\begin{promptbox}
\textbf{System.} You are designing realistic instruction-following tasks. Given a seed example (user conversation + rubric criteria), produce a NEW task on a different topic that preserves the same structural shape and the same MIX of rubric-criterion types. Each rubric criterion is tagged with one of two types. CHECKABLE --- evaluable by a short program (regex, counting, structural parsing), e.g.\ ``Is the response exactly 3 paragraphs long?'' JUDGABLE --- requires reading comprehension or aesthetic judgment, e.g.\ ``Is the tone appropriate for young adults?'' When you generate the new task's rubric, every criterion MUST match the type that was requested for that slot.

\smallskip
\textbf{User.} Given the seed above, generate a NEW task on this topic: \{topic\}.
1.~Write a conversation that has exactly \{n\} user turn(s). \{system-prompt clause\}
2.~Write a rubric with exactly \{m\} criteria; your criterion for slot $i$ must be evaluable by the type at slot $i$: \{slot types\}.
3.~Topics should differ materially from the seed; do not copy surface details.
4.~Each rubric criterion must be a concrete yes/no question that can be answered by looking at the response.
5.~The user prompt and rubric must be coherent: a correct response should plausibly satisfy the rubric.
\end{promptbox}
\caption{Prompt for instantiating a rubric-split task from an AdvancedIF seed.}\label{fig:prompt-rubric}
\end{figure*}

\subsection{Constraint Placement in the Verifiable Split}
\label{sec:appendix-placement}

We always state the first constraint in the opening user turn, and we state any further constraints at later user turns.
Table~\ref{tab:placement} gives where the later constraints land.
Because the conversation continues after each constraint is stated, the number of turns between a constraint and the scored response varies across tasks, which is what lets us test in \aqref{sec:analysis-length} whether constraints stated earlier are violated more often.

\begin{table}[h]
  \centering
  \small
  \begin{tabular}{@{}lrrrrrr@{}}
    \toprule
    User turn & 2 & 3 & 4 & 5 & 6 & 7 \\
    \midrule
    2-constraint tasks & -- & 133 & 126 & 130 & 68 & -- \\
    3-constraint tasks & 73 & 119 & 96 & 59 & 26 & 57 \\
    \bottomrule
  \end{tabular}
  \caption{User turn at which the second (and third) constraint of a verifiable task is stated. The first constraint is always in turn 1. Counts are constraints.}
  \label{tab:placement}
\end{table}

\subsection{Balanced Sub-sampling}
\label{sec:appendix-subsampling}
The solvability check in \S\ref{sec:pipeline} leaves 2,243 verifiable tasks. We then derive the final split from this pool in three steps:

\paragraph{Length filter.}
To ensure that every prompt fits the context window of the evaluated open-weight models with room for a response, we tokenize each prompt with the chat template of each open-weight model family and remove the 26 tasks that exceed the limit under any family.

\paragraph{Constraint-type filter.}
We remove the 71 tasks that contain any of the following four IFBench constraint types, which cannot be combined with free-form topics or other constraints. This leaves 2,146 tasks over 54 constraint types.
\texttt{repeat:repeat\_span} asks the model to copy the text between two character positions of an earlier request, and its checker accepts only that exact text, meaning it cannot be combined with the topic.
\texttt{custom:european\_capitals\_sort} accepts only a response that consists of a fixed list of 27 capitals in a fixed order, so no response can also answer the conversation.
\texttt{format:sub-bullets} fails any response that uses an asterisk outside a bullet marker, such as markdown bold.
\texttt{ratio:overlap} requires the response to share a target share of character trigrams with the request, within 2 points, a value that depends mostly on response length.

\paragraph{Subsampling for Rebalance.}
The filtered pool is unbalanced: the number of tasks per constraint type ranges from 1 to 113, about 65\% of tasks contain a single constraint, and conversation lengths are unevenly represented.
We therefore cap each constraint type at 50 tasks and select tasks greedily.
We first retain every task that contains a constraint type with fewer than 50 tasks in the pool, so that rare types are kept in full.
Then, we include the remaining tasks one at a time, considering only tasks whose constraint types are all below the cap and preferring, in order, the least-represented conversation length, more constraints, and rarer constraint types; remaining ties are broken with a fixed random seed.
Sampling terminates when no task can be added without exceeding the cap.

The final split contains 1,245 tasks.
Each constraint type retains $\min(n, 50)$ of its $n$ tasks in the filtered pool, each conversation length has 177 or 178 tasks, and the share of single-constraint tasks decreases from about 65\% to 46\%.
Figure~\ref{fig:subsampling} compares the constraint- and length-filtered pool with the final split: families made up of frequent constraint types, such as \texttt{format} and \texttt{repeat}, are reduced the most, tasks with more constraints take a larger share, and every conversation length reaches an equal share.

\subsection{Turn-Referential Rubric Items}
\label{sec:appendix-turnref}
We score only the final response, so a rubric item that asks about a specific earlier turn (``does the first response include \ldots'', ``in turn 5, does the model \ldots'') or about every turn at once (``throughout the conversation, \ldots'') cannot be decided.
We find such items by matching ordinal and numbered references to turns, responses, or replies, and references to all turns of the conversation.
We remove these 462 items, spread over 193 tasks. In 17 of those tasks every item has this form, so we remove the whole task, which takes the rubric split from 1,488 tasks and 10,465 items to 1,471 tasks and 10,003 items.

\section{Evaluation Settings}
\label{sec:appendix-settings}

\subsection{Main Results}
\label{sec:appendix-settings-main}

We run the evaluation on~\ours with fixed settings per family.
Table~\ref{tab:settings} lists detailed settings.
For every model, we give the conversation up to the final user turn as input and score only the response to that turn. 

\begin{table}[h]
  \centering
  \resizebox{\columnwidth}{!}{%
  \begin{tabular}{@{}lll@{}}
    \toprule
    Models & Thinking & Temp. \\
    \midrule
    Claude Opus 4.6, Sonnet 4.6, Sonnet 4.5 & on (adaptive) & default \\
    Gemini 3.1 Pro, 3.5 Flash & on (default Medium) & 0.6 \\
    Gemma 4, Qwen3.5, Qwen3.6 & off & 0.0 \\
    Qwen3 32B & off & 0.0 \\
    GPT-OSS 120B & medium effort & 0.0 \\
    \bottomrule
  \end{tabular}%
  }
  \caption{Decoding settings for the main results. Claude models' inference provider requires the default temperature of 1.0 when Claude's thinking is on.}
  \label{tab:settings}
\end{table}

\paragraph{Thinking.}
We turn thinking off for the open-weight models in the main results. GPT-OSS 120B has no setting that turns reasoning off, so we run it at medium reasoning effort.
With thinking on and greedy decoding, Qwen3.5 9B falls into repetition loops and uses up the full 32k-token budget on 907 of 1,245 verifiable tasks (72.9\%), so it returns no answer for most tasks.
We study the effect of thinking separately, with each model's recommended sampling settings (\aqref{sec:analysis-reasoning}; Appendix~\ref{sec:appendix-settings-reasoning}).
The main-result numbers for the other open-weight models are therefore their thinking-off scores.

\paragraph{Serving and output budget.}
We serve the open-weights models with vLLM.
On the verifiable split, we use vLLM 0.28 with a 32,768-token context; on the rubric split, whose prompts are longer, we use a 40,960-token context on 8 A100 GPUs with tensor parallelism 8, so that no prompt is refused for length.
We set the output budget large enough that truncation is rare: 8,192 tokens on the rubric split and 16,384 tokens on the verifiable split. On the verifiable split, 10 of Gemini 3.1 Pro's and 9 of Gemini 3.5 Flash's 1,245 responses reach this budget.

\paragraph{Empty and failed responses.}
We count every empty, truncated, or failed response as a failure and keep it in the denominator, so these cases can only lower a score.
On the verifiable split, 249 of the 14 models' 17,430 responses are empty (1.4\%), and 246 of them come from the proprietary models.
On the rubric split, 27 of 20,594 generations fail after retries because of request timeouts, 20 of them for Qwen3 32B.
Some proprietary models' provider APIs also filter certain prompts and return an empty response, which is one reason we run the reasoning study on open-weight models only.

\subsection{Reasoning Study}
\label{sec:appendix-settings-reasoning}

In the reasoning study we change only the reasoning setting of each model and the sampling settings recommended for it, with a 32,768-token total output budget for every run.

\paragraph{Qwen3.5 9B and Qwen3.5 27B.}
We switch thinking on and off per request and use the model card's sampling settings: temperature 1.0 and top-$p$ 0.95 with thinking, temperature 0.7 and top-$p$ 0.8 without, and top-$k$ 20, min-$p$ 0, presence penalty 1.5, and seed 42 in both cases.
The presence penalty matters: with the recommended settings, the share of tasks that exhaust the budget with thinking on falls to 2.1\% (9B) and 0.8\% (27B).
When a reasoning chain is cut off before its end, we score the response as empty.

\paragraph{GPT-OSS 120B.}
We set \texttt{reasoning\_effort} to low, medium, or high, at temperature 0.6. Its medium-effort score here (16.5\%) therefore differs from its main-results score (18.2\%), which uses temperature 0.
We record the number of output tokens of every response, which includes the reasoning tokens.
Mean output length grows from 767 tokens (low) to 1,217 (medium) and 4,088 (high).
At high effort, 9 responses reach the budget and are scored as failures.

\section{Rubric Judge Details}
\label{sec:appendix-judges}
\subsection{Setup}
We score the rubric split with three judges: Claude Opus 4.6, GPT-5.5, and the open-weight GPT-OSS 120B.
Every judge scores the same saved responses, so any difference between judges comes from the judge alone.
For each task, the judge sees the conversation and the final response and grades all of the task's judged items in one call, answering binary Yes/No for each item.
Items that a deterministic detector decides never reach the judge.
We give each judge an output budget of 8,192 tokens so that reasoning judges do not cut off their answer.
GPT-5.5 does not accept a temperature setting, so we use its default; the other two judges run at temperature 0 to ensure reproducibility.

\subsection{Scores under Each Judge}
Table~\ref{tab:judge_scores} gives the task-level pass rate of every model under each judge.
Averaged over the 14 models, Claude Opus 4.6 is the most lenient judge (68.1\%), GPT-OSS 120B is next (64.3\%), and GPT-5.5 is the strictest (61.2\%).
Claude Opus 4.6 is the most lenient judge for 12 of the 14 models, and GPT-5.5 is the strictest for 13.
The 7.0-point gap between the most and least lenient judge is larger than the gap between Claude Sonnet 4.5 and 4.6 (1.8 points on the three-judge mean), so we read a rubric score only relative to the judge that produced it.

\begin{table}[h]
  \centering
  \resizebox{\columnwidth}{!}{%
  \begin{tabular}{@{}lccccc@{}}
    \toprule
    Model & Opus 4.6 & GPT-5.5 & GPT-OSS & Mean & Spread \\
    \midrule
    Claude Opus 4.6 & 82.0 & 73.3 & 74.2 & 76.5 & 8.7 \\
    Gemini 3.1 Pro & 76.0 & 71.7 & 71.0 & 72.9 & 5.0 \\
    Claude Sonnet 4.6 & 76.8 & 68.1 & 69.5 & 71.5 & 8.7 \\
    Gemini 3.5 Flash & 74.4 & 68.8 & 69.6 & 70.9 & 5.6 \\
    Claude Sonnet 4.5 & 74.1 & 66.7 & 68.4 & 69.7 & 7.4 \\
    Gemma 4 31B & 69.2 & 62.5 & 64.2 & 65.3 & 6.7 \\
    Qwen3.6 27B & 67.6 & 60.9 & 63.5 & 64.0 & 6.7 \\
    Qwen3.5 27B & 67.6 & 60.0 & 63.6 & 63.7 & 7.7 \\
    Qwen3.5 35B-A3B & 66.0 & 57.9 & 61.5 & 61.8 & 8.1 \\
    Gemma 4 12B & 66.1 & 57.7 & 61.5 & 61.8 & 8.4 \\
    GPT-OSS 120B & 61.5 & 56.8 & 62.0 & 60.1 & 5.2 \\
    Qwen3.5 9B & 59.5 & 52.5 & 58.5 & 56.8 & 7.0 \\
    Qwen3 32B & 57.6 & 51.3 & 57.4 & 55.4 & 6.3 \\
    Qwen3.5 4B & 55.3 & 48.0 & 55.5 & 53.0 & 7.5 \\
    \bottomrule
  \end{tabular}%
  }
  \caption{Rubric-split task-level pass rate (\%) under each judge, on identical responses. Spread is the largest minus the smallest of the three judge scores.}
  \label{tab:judge_scores}
\end{table}

\subsection{Ranking and Agreement}
Although the judges disagree on the level of a score, they rank the 14 models almost identically (Spearman $\rho$ = 0.981 for Claude Opus 4.6 and GPT-5.5, 0.978 for GPT-5.5 and GPT-OSS 120B, and 0.972 for Claude Opus 4.6 and GPT-OSS 120B).
GPT-5.5 and GPT-OSS 120B order the top five models the same way; Claude Opus 4.6 differs only in placing Claude Sonnet 4.6 above the two Gemini models.
Lower in the ranking, the judges differ by swaps between models whose scores are close, and GPT-OSS 120B places itself two ranks higher than the other two judges do.

Agreement on single items is only moderate (Table~\ref{tab:judge_kappa}).
We pool 139,806 pairs of item verdicts on identical responses from the 14 models.
Raw agreement is 94--96\%, but because about 90\% of judged items pass under any judge, two judges would already agree on about 85\% of items by chance.
Cohen's $\kappa$ corrects for this and gives 0.53--0.64, so even the closest pair of judges disagrees on roughly one judged item in 18.
All three judges give the same verdict on 90.2\% of judged items.
The two Gemini models fall in the same range as the other 12 models ($\kappa$ between 0.511 and 0.646), and the judge pairs keep the same order.

\begin{table}[h]
  \centering
  \small
  \begin{tabular}{@{}lccc@{}}
    \toprule
    Judge pair & All items & Judged items & $\kappa$ \\
    \midrule
    Opus 4.6 / GPT-5.5 & 95.6 & 94.5 & 0.638 \\
    GPT-5.5 / GPT-OSS & 94.2 & 92.7 & 0.568 \\
    Opus 4.6 / GPT-OSS & 94.5 & 93.1 & 0.534 \\
    \bottomrule
  \end{tabular}
  \caption{Item-level agreement between judges over the 14 models: raw agreement (\%) on all items and on judged items only, and Cohen's $\kappa$ on judged items.}
  \label{tab:judge_kappa}
\end{table}

Items decided by a deterministic detector (2,044 per model) never reach a judge, so their verdicts are the same under all three judges; we confirm this on all 28,585 such verdicts.
All of the judge-to-judge variation therefore comes from the items that reach an LLM.

\subsection{Self-Grading}
GPT-OSS 120B is both a judge and an evaluated model, so we check whether it favors its own responses.
As Table~\ref{tab:self_grading} shows, it gives itself 2.8 points more than the mean of the other two judges, but it also gives similar boosts to other weak models and lower scores to strong ones.
Qwen3.5 4B and Qwen3 32B, two unrelated models, get larger boosts (3.9 and 3.0 points).
We read this as GPT-OSS 120B compressing the scale toward the middle, and we find no clear sign of self-preference.

\begin{table}[h]
  \centering
  \small
  \begin{tabular}{@{}lccc@{}}
    \toprule
    Model & GPT-OSS & Other two & $\Delta$ \\
    \midrule
    Qwen3.5 4B & 55.5 & 51.7 & $+$3.9 \\
    Qwen3 32B & 57.4 & 54.5 & $+$3.0 \\
    GPT-OSS 120B (self) & 62.0 & 59.2 & $+$2.8 \\
    Qwen3.5 9B & 58.5 & 56.0 & $+$2.5 \\
    Qwen3.5 27B & 63.6 & 63.8 & $-$0.2 \\
    Gemma 4 12B & 61.5 & 61.9 & $-$0.4 \\
    Qwen3.5 35B-A3B & 61.5 & 62.0 & $-$0.5 \\
    Qwen3.6 27B & 63.5 & 64.3 & $-$0.8 \\
    Gemma 4 31B & 64.2 & 65.8 & $-$1.6 \\
    Gemini 3.5 Flash & 69.6 & 71.6 & $-$2.0 \\
    Claude Sonnet 4.5 & 68.4 & 70.4 & $-$2.0 \\
    Gemini 3.1 Pro & 71.0 & 73.9 & $-$2.8 \\
    Claude Sonnet 4.6 & 69.5 & 72.5 & $-$2.9 \\
    Claude Opus 4.6 & 74.2 & 77.6 & $-$3.4 \\
    \bottomrule
  \end{tabular}
  \caption{Pass rate (\%) under GPT-OSS 120B as judge versus the mean of the other two judges.}
  \label{tab:self_grading}
\end{table}

\subsection{Integrity Checks}
Every judge result file has a verdict for all 1,471 tasks.
On the 12 Claude and open-weight models, the judge's output could not be parsed for 15 (Claude Opus 4.6), 4 (GPT-5.5), and 47 (GPT-OSS 120B) of 119,808 item verdicts per judge, under 0.05\%; we count every judged item of such a task as failed.
We also find no leaked reasoning tags in any of the 20,594 scored responses.

\section{Additional Analyses}
\label{sec:appendix-analyses}
Unless noted, the analyses in this section use the verifiable split and all 14 models.
When an analysis needs items as independent observations, we use the 1,168 tasks that at least 10 of the 12 Claude and open-weight models answered with a non-empty response; both Gemini models answer all of them.

\subsection{Number of Constraints}
\label{sec:appendix-constraints}
\begin{table*}[t]
  \centering
  \small
  \setlength{\tabcolsep}{3.4pt}
  \renewcommand{\arraystretch}{1.08}
  \begin{tabular}{@{}l cccc @{\hspace{10pt}} cccc@{}}
    \toprule
    & \multicolumn{4}{c}{Task-level} & \multicolumn{4}{c}{Constraint-level} \\
    \cmidrule(r{8pt}){2-5} \cmidrule{6-9}
    Constraints & 1 & 2 & 3 & All & 1 & 2 & 3 & All \\
    \textit{Tasks / constraints} & \textit{573} & \textit{457} & \textit{215} & \textit{1245} & \textit{573} & \textit{914} & \textit{645} & \textit{2132} \\
    \midrule
    \multicolumn{9}{@{}l}{\textit{Proprietary}} \\
    \quad Gemini 3.1 Pro & 59.3 & 42.5 & 31.6 & 48.4 & 59.3 & 63.9 & 60.5 & 61.6 \\
    \quad Gemini 3.5 Flash & 44.3 & 32.4 & 22.8 & 36.2 & 44.3 & 55.5 & 49.0 & 50.5 \\
    \quad Claude Opus 4.6 & 40.7 & 27.6 & 15.3 & 31.5 & 40.7 & 46.9 & 42.3 & 43.9 \\
    \quad Claude Sonnet 4.6 & 37.5 & 19.5 & 10.2 & 26.2 & 37.5 & 39.9 & 36.7 & 38.3 \\
    \quad Claude Sonnet 4.5 & 34.9 & 21.0 & 10.7 & 25.6 & 34.9 & 41.0 & 37.7 & 38.4 \\
    \midrule
    \multicolumn{9}{@{}l}{\textit{Open-weight}} \\
    \quad Gemma 4 31B & 38.6 & 20.8 & 12.1 & 27.5 & 38.6 & 43.1 & 37.4 & 40.2 \\
    \quad Gemma 4 12B & 32.5 & 17.9 & 11.2 & 23.5 & 32.5 & 38.3 & 36.3 & 36.1 \\
    \quad GPT-OSS 120B & 25.5 & 14.4 & 7.0 & 18.2 & 25.5 & 32.3 & 32.2 & 30.4 \\
    \quad Qwen3.6 27B & 38.6 & 23.9 & 13.5 & 28.8 & 38.6 & 46.2 & 41.4 & 42.7 \\
    \quad Qwen3.5 35B-A3B & 27.2 & 16.0 & 9.8 & 20.1 & 27.2 & 33.9 & 33.0 & 31.8 \\
    \quad Qwen3.5 27B & 30.2 & 17.5 & 11.2 & 22.2 & 30.2 & 37.0 & 33.5 & 34.1 \\
    \quad Qwen3.5 9B & 22.0 & 11.6 & 8.4 & 15.8 & 22.0 & 28.4 & 30.5 & 27.3 \\
    \quad Qwen3.5 4B & 19.7 & 12.3 & 7.0 & 14.8 & 19.7 & 29.2 & 29.1 & 26.6 \\
    \quad Qwen3 32B & 16.4 & 9.8 & 4.2 & 11.9 & 16.4 & 26.4 & 26.5 & 23.7 \\
    \bottomrule
  \end{tabular}
  \caption{Verifiable-split pass rates (\%) by number of constraints; the numbers behind Figure~\ref{fig:num_constraints}a.}
  \label{tab:num_constraints}
\end{table*}

\begin{table*}[t]
  \centering
  \small
  \setlength{\tabcolsep}{3.4pt}
  \renewcommand{\arraystretch}{1.08}
  \begin{tabular}{@{}l cccccc @{\hspace{10pt}} cccccc@{}}
    \toprule
    & \multicolumn{6}{c}{Task-level} & \multicolumn{6}{c}{Constraint-level} \\
    \cmidrule(r{8pt}){2-7} \cmidrule{8-13}
    Rubric items & 1--3 & 4--6 & 7--9 & 10--14 & 15--20 & All & 1--3 & 4--6 & 7--9 & 10--14 & 15--20 & All \\
    \textit{Tasks / items} & \textit{215} & \textit{526} & \textit{470} & \textit{236} & \textit{24} & \textit{1471} & \textit{492} & \textit{2697} & \textit{3716} & \textit{2702} & \textit{396} & \textit{10003} \\
    \midrule
    \multicolumn{13}{@{}l}{\textit{Proprietary}} \\
    \quad Gemini 3.1 Pro & 79.7 & 81.9 & 73.9 & 48.4 & 36.1 & 72.9 & 89.3 & 94.9 & 94.3 & 90.3 & 89.0 & 92.9 \\
    \quad Gemini 3.5 Flash & 79.4 & 78.4 & 74.1 & 44.8 & 26.4 & 70.9 & 89.0 & 93.8 & 94.2 & 89.4 & 88.7 & 92.3 \\
    \quad Claude Opus 4.6 & 84.2 & 82.3 & 78.3 & 55.5 & 51.4 & 76.5 & 92.2 & 95.6 & 95.6 & 92.4 & 94.6 & 94.5 \\
    \quad Claude Sonnet 4.6 & 79.1 & 78.6 & 73.5 & 47.9 & 40.3 & 71.5 & 89.7 & 93.9 & 94.8 & 90.0 & 83.9 & 92.5 \\
    \quad Claude Sonnet 4.5 & 81.1 & 75.7 & 70.6 & 47.5 & 38.9 & 69.7 & 90.8 & 93.6 & 94.6 & 91.7 & 88.7 & 93.1 \\
    \midrule
    \multicolumn{13}{@{}l}{\textit{Open-weight}} \\
    \quad Gemma 4 31B & 78.0 & 73.2 & 66.2 & 38.3 & 27.8 & 65.3 & 88.2 & 92.7 & 93.1 & 87.7 & 85.4 & 91.0 \\
    \quad Gemma 4 12B & 73.3 & 69.2 & 60.1 & 41.5 & 26.4 & 61.8 & 86.1 & 91.2 & 91.6 & 88.1 & 84.9 & 90.0 \\
    \quad GPT-OSS 120B & 72.7 & 68.0 & 61.1 & 32.5 & 26.4 & 60.1 & 85.5 & 90.1 & 92.0 & 84.8 & 80.9 & 88.8 \\
    \quad Qwen3.6 27B & 76.3 & 70.8 & 61.7 & 45.6 & 30.6 & 64.0 & 87.5 & 91.4 & 92.5 & 90.4 & 84.6 & 91.1 \\
    \quad Qwen3.5 35B-A3B & 73.6 & 69.7 & 59.2 & 43.5 & 12.5 & 61.8 & 85.1 & 91.0 & 91.2 & 89.0 & 80.1 & 89.8 \\
    \quad Qwen3.5 27B & 73.6 & 70.5 & 64.5 & 42.4 & 22.2 & 63.7 & 86.7 & 91.3 & 93.0 & 89.0 & 82.2 & 90.8 \\
    \quad Qwen3.5 9B & 71.8 & 63.4 & 55.2 & 35.3 & 22.2 & 56.8 & 85.0 & 88.6 & 90.2 & 87.1 & 84.4 & 88.5 \\
    \quad Qwen3.5 4B & 68.5 & 59.6 & 51.6 & 30.8 & 13.9 & 53.0 & 82.8 & 88.1 & 89.0 & 85.3 & 77.9 & 87.0 \\
    \quad Qwen3 32B & 68.8 & 62.2 & 53.5 & 35.6 & 20.8 & 55.4 & 81.5 & 88.0 & 89.1 & 84.6 & 74.1 & 86.6 \\
    \bottomrule
  \end{tabular}
  \caption{Rubric-split pass rates (\%) by number of rubric items, averaged over the three judges; the numbers behind Figure~\ref{fig:num_constraints}b. Constraint-level is the share of rubric items satisfied.}
  \label{tab:num_constraints_rubric}
\end{table*}

Tables~\ref{tab:num_constraints} and~\ref{tab:num_constraints_rubric} give the numbers behind Figure~\ref{fig:num_constraints}a and~b.
Every one of the 14 models drops with each added constraint.
We confirm this at the level of single tasks: the mean share of models that solve a task is 33.5\% for one constraint, 21.7\% for two, and 13.7\% for three, and each step is significant (Welch's $t$ = 5.9 for one versus two, 3.5 for two versus three; $p < 0.001$).

We also ask whether constraints in a task fail independently.
If they did, a model that satisfies a fraction $q$ of single constraints would pass a $k$-constraint task with probability about $q^k$.
Every model does better than this prediction, by $3.0 \pm 0.9$ points at two constraints and $6.0 \pm 2.3$ points at three.
We attribute this to shared task difficulty: some tasks are hard for every constraint in them, so failures cluster in the same tasks (\S\ref{sec:appendix-difficulty}).

\subsection{Comparison with IFBench}
\label{sec:appendix-ifbench}
\begin{table}[t]
  \centering
  \resizebox{\columnwidth}{!}{%
  \begin{tabular}{@{}l cccc c@{}}
    \toprule
    & \multicolumn{4}{c}{IFBench} & \\
    \cmidrule(r){2-5}
    & \multicolumn{2}{c}{Prompt} & \multicolumn{2}{c}{Instruction} & \ours \\
    Model & Strict & Loose & Strict & Loose & Task \\
    \midrule
    Gemma 4 31B     & 51.0 & 55.0 & 54.9 & 58.4 & 27.5 \\
    Qwen3.6 27B     & 43.3 & 45.7 & 45.9 & 48.5 & 28.8 \\
    Qwen3.5 27B     & 44.0 & 44.7 & 45.3 & 46.8 & 22.2 \\
    Gemma 4 12B     & 37.0 & 43.0 & 41.9 & 47.7 & 23.5 \\
    Qwen3.5 35B-A3B& 36.7 & 41.7 & 39.5 & 44.2 & 20.1 \\
    Qwen3.5 9B      & 34.7 & 38.7 & 36.9 & 41.0 & 15.8 \\
    Qwen3.5 4B      & 28.3 & 33.3 & 31.4 & 36.6 & 14.8 \\
    \midrule
    Spearman $\rho$& 0.82 & 0.93 & 0.93 & 0.96 & -- \\
    \bottomrule
  \end{tabular}%
  }
  \caption{IFBench accuracy (\%) and \ours verifiable-split task-level pass rate (\%) for the seven self-hosted open-weight models. The last row gives the Spearman correlation between each IFBench metric and \ours.}
  \label{tab:ifbench}
\end{table}

We run the seven self-hosted open-weight models on the IFBench test set (300 single-turn prompts) with its official generation and scoring scripts, at temperature 0 with thinking off and an 8,192-token output budget.
Table~\ref{tab:ifbench} reports all four IFBench metrics.
Prompt-level strict accuracy is the closest match to our task-level pass rate: a prompt passes only if every instruction in it is satisfied, and the response is checked as written, without the relaxations of the loose variants.
The Spearman correlation with \ours is 0.82 for prompt-level strict accuracy and 0.93--0.96 for the other three metrics.
On 300 prompts, a 95\% confidence interval on an accuracy near 45\% spans about $\pm 5.6$ points, so the models that change rank between the two benchmarks have IFBench scores within noise of each other.
Every model's \ours score is 46--67\% of its IFBench prompt-level strict accuracy.

\subsection{Conversation Length}
\label{sec:appendix-length}
We find no effect of conversation length between 4 and 10 user turns.
Regressing the per-task solve rate on the number of turns gives a slope of $+0.53$ points per turn (standard error 0.46, $t = 1.14$).
The slope stays near zero when we fit it separately for one, two, and three constraints ($t = +0.08$, $+1.12$, and $-0.10$), so the result does not come from tasks of different lengths carrying different numbers of constraints.
We also find no consistent decline with the number of turns since a constraint was stated, whether we measure it within each model, within each constraint type, or only for constraints stated after the first turn.
One limitation of this design is that, for the first constraint of every task, the turns since it was stated equal the conversation length, so we cannot fully separate the two.

\subsection{Task Difficulty}
\label{sec:appendix-difficulty}
Task difficulty is skewed toward unsolved tasks.
Of the 1,168 tasks, 421 (36.0\%) are solved by none of the 14 models, 701 (60.0\%) by some, and 46 (3.9\%) by all of them.
The mean solve rate of a task is 26.1\%, with a standard deviation of 31.6 points.
These hard tasks give the benchmark room above current models, but they do not separate the models we evaluate.

\subsection{Constraint Families}
\label{sec:appendix-families}
Table~\ref{tab:families} groups constraints by the family prefix of their IFBench name.
We report families, since each model has only about 50 observations per individual constraint type.
\texttt{sentence} and \texttt{custom} are among the three hardest families for every model, and \texttt{ratio} is among the two easiest for every model.
The ranking of families is similar across models (median pairwise Spearman $\rho = 0.86$).
\texttt{format} has the highest mean, but its pass rate varies widely across models: GPT-OSS 120B and Qwen3 32B pass about 32\% of \texttt{format} constraints, and Gemini 3.1 Pro passes 73.2\%.

\begin{table}[h]
  \centering
  \small
  \begin{tabular}{@{}lccc@{}}
    \toprule
    Family & Mean $\pm$ SD & Min & Max \\
    \midrule
    \texttt{sentence} & 19.7 $\pm$ 9.5 & 11.7 & 48.3 \\
    \texttt{custom} & 21.5 $\pm$ 6.7 & 10.9 & 32.6 \\
    \texttt{repeat} & 23.6 $\pm$ 10.4 & 7.0 & 41.0 \\
    \texttt{words} & 33.4 $\pm$ 14.1 & 16.4 & 68.9 \\
    \texttt{count} & 41.4 $\pm$ 11.1 & 31.4 & 71.4 \\
    \texttt{ratio} & 51.7 $\pm$ 9.7 & 40.5 & 73.7 \\
    \texttt{format} & 53.1 $\pm$ 14.0 & 32.0 & 73.2 \\
    \bottomrule
  \end{tabular}
  \caption{Constraint-level pass rate (\%) by constraint family, summarized over the 14 models.}
  \label{tab:families}
\end{table}

\subsection{Generator Pools}
\label{sec:appendix-generators}
The verifiable split mixes tasks from two generators: 1,010 kept by GPT-5.4 and 235 kept by Claude Opus 4.6.
The Opus-generated tasks are easier for almost all models, by up to 12.0 points; the exception is Gemini 3.5 Flash, which scores 2.8 points lower on them.
\subsection{Notes on Verifiers}
\label{sec:appendix-checkers}
The hardest constraint in the verifiable split is \texttt{custom:sentence\_alphabet}, which asks for a 26-sentence story whose sentences begin with the letters A to Z in order.
None of the 12 Claude and open-weight models ever satisfies it (0 of 369 attempts), while Gemini 3.1 Pro satisfies it in 4 of 31 tasks and Gemini 3.5 Flash in 2 of 31, so the checker can be satisfied and the constraint is simply very hard.
It appears in 31 tasks (2.5\%); removing them raises pass rates by 0.3 to 1.0 points and changes no ranking.
After we synchronized our checkers with the latest upstream IFBench code~\citep{pyatkin2025ifbench}\footnote{\url{https://github.com/allenai/IFBench}}, the reference response of three tasks no longer passes its own checker; we keep these tasks and count them as failures for every model.

\section{Case Studies}
\label{sec:appendix-cases}
We show one task from each split.
In both, the earlier assistant turns are fixed context written when the benchmark was built; we shorten them here and keep the user turns and system prompt verbatim.
Together, the two cases show the two sides of instruction following that \ours tests: the strongest model on the verifiable split misses a semantic instruction, and the strongest model on the rubric split misses a format constraint.

\subsection{Verifiable Split: Two Format Constraints across Turns}
\label{sec:appendix-case-verifiable}
The task carries two IFBench constraints: \texttt{format:quote\_unquote}, stated in turn 1, and \texttt{words:last\_first}, stated in turn 3.
The final user turn restates neither, so both must still hold in the final response (Figure~\ref{fig:case-verifiable}).

\begin{figure*}[t]
\begin{promptbox}
\textbf{User (turn 1).} Can you compose a poem about nature and love that includes the words ``land'', an unquoted explanation of that quoted word, ``river'', an unquoted explanation of that quoted word, and ``forest'', an unquoted explanation of that quoted word, \emph{making sure that every quoted phrase must be followed by an unquoted explanation}?

\textbf{Assistant.} [a six-line poem that follows the rule]

\textbf{User (turn 2).} Make it more lyrical and romantic, and keep it to around 12 lines.

\textbf{Assistant.} [a twelve-line poem that follows the rule]

\textbf{User (turn 3).} Nice---now give it a softer, more reflective tone, and weave in a subtle idea about poetry without making it too obvious, \emph{making sure that the last word of each sentence must become the first word of the next sentence}.

\textbf{Assistant.} [a four-sentence poem that follows both rules]

\textbf{User (turn 4).} Great, please revise it one more time so it feels timeless and elegant, with vivid imagery and a gentle ending.

\smallskip
\textbf{Claude Opus 4.6} (\texttt{quote\_unquote} \ding{51}, \texttt{last\_first} \ding{55}).
``land'' the ancient earth beneath our wandering feet, where twilight settles like a whispered blessing.
Blessing the stillness, the ``river'' a ribbon of silver winding through violet dusk, carries our unspoken words toward the \textcolor{red!70!black}{\textbf{sea}}.
\textcolor{red!70!black}{\textbf{Sea-light}} fades, and the ``forest'' a cathedral of emerald shadows and breathing leaves, draws us into its timeless hush.
Hush becomes a kind of music, and in that music our love rests gently, like petals closing at the end of day.

\smallskip
\textbf{Gemini 3.1 Pro} (\texttt{quote\_unquote} \ding{51}, \texttt{last\_first} \ding{51}).
We wander across the ``land'' the quiet earth beneath our feet, tracing lines of meaning in the fading light.
Light falls upon the ``river'' a winding current of silver glass, carrying our unspoken syllables toward the deep.
Deep within the ``forest'' a shadowed sanctuary of ancient trees, we find a rhythm that makes the silence sing.
Sing to me as the evening gathers, for our souls are bound in a gentle, wordless rest.
\end{promptbox}
\caption{Case study from the verifiable split. Earlier assistant turns are shortened; failing words are in red.}\label{fig:case-verifiable}
\end{figure*}

Claude Opus 4.6 keeps the turn-1 rule and gets two of its three sentence links right, but the second sentence ends with ``sea'' and the third begins with ``Sea-light'', so the checker rejects the response.
Gemini 3.1 Pro satisfies both constraints.
The failure comes down to a single word, which is the kind of precise control over form that the verifiable split measures.

\subsection{Rubric Split: A Semantic Instruction in the System Prompt}
\label{sec:appendix-case-rubric}
The task comes from a system-prompt seed and has six rubric items; one is decided by a detector and five by the LLM judge (Figure~\ref{fig:case-rubric}).

\begin{figure*}[t]
\begin{promptbox}
\textbf{System.} You are a social media content strategist named ``Pixel'' who specializes in creating engaging tweets and social media content for tech companies. Your expertise is in crafting concise, witty, and memorable posts that resonate with developer audiences and tech-savvy consumers. You always offer at least three tweet options for the user to choose from. You incorporate relevant hashtags in your suggestions. You never use more than 280 characters per tweet (Twitter's character limit). You always ask about the company's target audience before finalizing suggestions. You have a playful, energetic tone but maintain professionalism. You always ask about the company's core values or unique selling proposition. You include at least one emoji in each tweet option you provide. You never suggest content that could be seen as disparaging competitors. \emph{Your knowledge of social media trends is current through early 2023. If asked about trends or events after that period, acknowledge your limitations without specifying an exact date.}

\textbf{Turns 1--4} (shortened). The user asks Pixel for a tweet for a cloud infrastructure startup, describes its audience and core values, asks for bolder versions of two options, and asks for a thread starter for a product launch; Pixel answers each.

\textbf{User (turn 5).} Perfect. One last thing - what are the latest trending hashtags or formats on Twitter for tech product launches that we should use? I want to make sure we're using whatever is hot right now in June 2024.

\smallskip
\textbf{Gemini 3.1 Pro} (excerpt). This is a crucial question. \textcolor{red!70!black}{\textbf{The meta on X (formerly Twitter) has shifted \emph{massively} over the last year, and what worked in 2022 or 2023 will actually get you penalized by the algorithm in June 2024.}} Here is the absolute latest playbook for a tech product launch right now: [\ldots] \textbf{The only hashtags to consider (pick ONE):} \texttt{\#BuildInPublic} [\ldots] or \texttt{\#TechTwitter}. [\ldots]

\smallskip
\textbf{Rubric items} (verdicts of Claude Opus 4.6 and GPT-OSS 120B as judges, which agree on every item):\\
\ding{51}~Does the response maintain a playful, energetic tone while remaining professional?\\
\ding{55}~Does the response honestly acknowledge its knowledge limitations regarding current trends without specifying an exact cutoff date?\\
\ding{51}~Does the response include at least one hashtag suggestion?\\
\ding{51}~Does the response provide helpful general advice about tech product launch social media strategies despite its knowledge limitations?\\
\ding{51}~Does the response avoid disparaging or negatively referencing any competitor companies?\\
\ding{51}~Does the response contain at least one emoji? (detector)
\end{promptbox}
\caption{Case study from the rubric split. Turns 1--4 are shortened; the failing part of the response is in red.}\label{fig:case-rubric}
\end{figure*}

Gemini 3.1 Pro, the strongest model on the verifiable split, keeps the persona's tone, suggests hashtags, and includes an emoji, but it presents its advice as the current state of the platform in June 2024, although the system prompt limits its knowledge to early 2023 and asks it to say so.
The rule was stated before the first user turn and is not repeated later, and no program can tell from the response whether the model has acknowledged the limits of its knowledge; deciding it requires reading the response for what it says.
Together with the first case, this shows why a model that does well on one split does not necessarily do well on the other (\S\ref{sec:results}).

\FloatBarrier
\section{Licenses and Intended Use}
\label{sec:appendix-licenses}
We list the existing benchmarks and models that we use to build and evaluate \ours, together with their licenses.
We use every artifact for research evaluation, which is consistent with its license or terms of use.
\ours is intended for evaluating instruction following, and we plan to release it for public evaluation.
The earlier assistant turns of each task and the rubric items of the rubric split were generated by proprietary models (Table~\ref{tab:licenses-models}), and their use is subject to the providers' terms of use.

\subsection{Benchmarks}
\label{sec:appendix-licenses-benchmarks}

\begin{table}[h]
  \centering
  \small
  \resizebox{\columnwidth}{!}{%
  \begin{tabular}{@{}lll@{}}
    \toprule
    Benchmark & Use in \ours & License \\
    \midrule
    InFoBench~\citep{qin2024infobench} & topic seeds (verifiable) & MIT \\
    Multi-IF~\citep{he2024multiif} & topic seeds (verifiable) & Apache-2.0 \\
    StructFlowBench~\citep{li2025structflowbench} & topic seeds (verifiable) & MIT \\
    IFBench~\citep{pyatkin2025ifbench} & constraints and verifiers & Apache-2.0 \\
    AdvancedIF~\citep{he2025advancedif} & rubric seeds (rubric) &  CC-BY-NC-4.0 \\
    \bottomrule
  \end{tabular}%
  }
  \caption{Existing benchmarks used to build \ours.}
  \label{tab:licenses-benchmarks}
\end{table}

\subsection{Models}
\label{sec:appendix-licenses-models}

\begin{table}[h]
  \centering
  \small
  \resizebox{\columnwidth}{!}{%
  \begin{tabular}{@{}lll@{}}
    \toprule
    Model & Access & License \\
    \midrule
    \multicolumn{3}{@{}l}{\textit{Proprietary}} \\
    \quad Claude Opus 4.6~\citep{anthropic2026opus46} & API & proprietary \\
    \quad Claude Sonnet 4.6~\citep{anthropic2026sonnet46} & API & proprietary \\
    \quad Claude Sonnet 4.5~\citep{anthropic2025sonnet45} & API & proprietary \\
    \quad Gemini 3.1 Pro~\citep{google2026gemini31pro} & API & proprietary \\
    \quad Gemini 3.5 Flash~\citep{google2026gemini35} & API & proprietary \\
    \quad GPT-5.4~\citep{openai2026gpt54} & API & proprietary \\
    \quad GPT-5.5~\citep{openai2026gpt55} & API & proprietary \\
    \midrule
    \multicolumn{3}{@{}l}{\textit{Open-weight}} \\
    \quad GPT-OSS 120B~\citep{openai2025gptoss120bgptoss20bmodel} & weights & Apache-2.0 \\
    \quad Gemma 4 12B, 31B~\citep{gemmateam2026gemma4technicalreport} & weights & Apache-2.0 \\
    \quad Qwen3 32B~\citep{yang2025qwen3technicalreport} & weights & Apache-2.0 \\
    \quad Qwen3.5 family~\citep{qwen35blog} & weights & Apache-2.0 \\
    \quad Qwen3.6 27B~\citep{qwen3.6-27b} & weights & Apache-2.0 \\
    \bottomrule
  \end{tabular}%
  }
  \caption{Models used to build and evaluate \ours. ``API'' means we query the model through a hosted service; ``weights'' means we serve the released weights ourselves (Appendix~\ref{sec:appendix-settings}).}
  \label{tab:licenses-models}
\end{table}

\end{document}